\pdfoutput=1
\documentclass{article}

\usepackage{arxiv}

\usepackage[utf8]{inputenc}
\usepackage[T1]{fontenc}
\usepackage{hyperref}
\usepackage{url}
\usepackage[dvipsnames]{xcolor}
\usepackage{graphicx}
\usepackage{subcaption}
\usepackage{array}
\usepackage{booktabs}
\usepackage{natbib}
\usepackage{microtype}
\usepackage{doi}
\usepackage{listings}
\usepackage{xcolor}
\usepackage{subcaption}

\definecolor{codebg}{HTML}{F7F7F5}
\definecolor{codeframe}{HTML}{E2E5E9}
\definecolor{codetext}{HTML}{2F343D}
\definecolor{codenum}{HTML}{A7ADB5}

\definecolor{codekeyword}{HTML}{6E6CCF}  
\definecolor{codecomment}{HTML}{A6ADB8}  
\definecolor{codestring}{HTML}{B38AD8}   
\definecolor{codefunc}{HTML}{B46A7A}     
\definecolor{codeconst}{HTML}{5E9C93}    

\lstdefinestyle{acm_style}{
    backgroundcolor=\color{codebg},
    basicstyle=\ttfamily\footnotesize\color{codetext},
    commentstyle=\color{codecomment},
    keywordstyle=\bfseries\color{codekeyword},
    stringstyle=\color{codestring},
    identifierstyle=\color{codetext},
    numberstyle=\scriptsize\color{codenum},
    numbers=left,
    numbersep=8pt,
    xleftmargin=4pt,
    frame=single,
    rulecolor=\color{codeframe},
    framerule=0.4pt,
    framesep=5pt,
    breaklines=true,
    breakatwhitespace=false,
    showspaces=false,
    showstringspaces=false,
    showtabs=false,
    keepspaces=true,
    tabsize=4,
    captionpos=b,
    aboveskip=6pt,
    belowskip=6pt,
    emph={range,len,min,max,sum,zip,enumerate,print,abs,sorted,np,array,zeros,ones,clip,tile,uniform},
    emphstyle=\color{codefunc},
    emph={[2]True,False,None},
    emphstyle={[2]\color{codeconst}}
}

\usepackage{amsmath,amsfonts,bm}

\def\eqref#1{equation~\ref{#1}}

\def\1{\bm{1}}

\DeclareMathAlphabet{\mathsfit}{\encodingdefault}{\sfdefault}{m}{sl}
\SetMathAlphabet{\mathsfit}{bold}{\encodingdefault}{\sfdefault}{bx}{n}

\usepackage[ruled,vlined]{algorithm2e}

\SetCommentSty{mycommfont}
\SetKwComment{Comment}{// }{}
\setcitestyle{authoryear,round,citesep={;},aysep={,},yysep={;}}

\title{Escher-Loop: Mutual Evolution by Closed-Loop Self-Referential Optimization}

\author{
    Ziyang Liu\textsuperscript{\dag} \\
        Shenzhen X-Institute; \\
        Soochow University \\
        \texttt{newzil1225@gmail.com}
    \And
    Xinyan Guo\textsuperscript{\dag} \\
        Shenzhen X-Institute; \\
        Shenzhen Loop Area Institute \\
        \texttt{guoxinyan@mails.x-institute.edu.cn}
    \And
    Xuchen Wei\textsuperscript{\dag} \\ 
        Shenzhen X-Institute \\
        \texttt{weixuchen@mails.x-institute.edu.cn}
    \And
    Han Hao\textsuperscript{*} \\
        Tsinghua University\\
        \texttt{haoh23@mails.tsinghua.edu.cn}
    \And
    Liu Yang\textsuperscript{*} \\ 
        National University of Singapore\\
        \texttt{yangliu@nus.edu.sg}
}

\renewcommand{\shorttitle}{Escher-Loop}
\date{}

\hypersetup{
pdftitle={Escher-Loop: Mutual Evolution by Closed-Loop Self-Referential Optimization},
}

\begin{document}
\maketitle

\begingroup
\renewcommand\thefootnote{}\footnotetext{%
	\textsuperscript{\dag}\;Equal contribution; \quad
	\textsuperscript{*}\;Corresponding author.
}
\addtocounter{footnote}{-1}
\endgroup

\begin{abstract}

While recent autonomous agents demonstrate impressive capabilities, they predominantly rely on manually scripted workflows and handcrafted heuristics, inherently limiting their potential for open-ended improvement. To address this, we propose \emph{Escher-Loop}, a fully closed-loop framework that operationalizes the mutual evolution of two distinct populations: \emph{Task Agents} that solve concrete problems, and \emph{Optimizer Agents} that recursively refine both the task agents and themselves. To sustain this self-referential evolution, we propose a dynamic benchmarking mechanism that seamlessly reuses the empirical scores of newly generated task agents as relative win-loss signals to update optimizers' scores. This mechanism leverages the evolution of task agents as an inherent signal to drive the evaluation and refinement of optimizers without additional overhead. Empirical evaluations on mathematical optimization problems demonstrate that Escher-Loop effectively pushes past the performance ceilings of static baselines, achieving the highest absolute peak performance across all evaluated tasks under matched compute. Remarkably, we observe that the optimizer agents dynamically adapt their strategies to match the shifting demands of high-performing task agents, which explains the system's continuous improvement and superior late-stage performance.
\end{abstract}

\section{Introduction}
\label{sec:introduction}

Despite the rapid development of autonomous agents \citep{yang2023autogptonlinedecisionmaking,anthropic_claudecode,luo2025largelanguagemodelagent}, the majority of current systems remain limited to human-designed tools and manually scripted workflows \citep{yao2023react,shinn_reflexion_2023}. These systems are often engineered for specific tasks, relying heavily on human priors to navigate complex problem spaces \citep{jimenez2024swebench,wang2024voyager,yang_swe-agent_2024}. While effective in narrow domains, this reliance on handcrafted heuristics inherently limits their generalization and caps their potential for autonomous growth. This trend stands in direct opposition to the ``bitter lesson'' \citep{sutton2019bitter}: \emph{history has shown that methods leveraging scalable computation and adaptive search invariably outperform those built on human-engineered intuition.} While the pursuit of recursive self-improvement to overcome these limits has a rich history \citep{good1966speculations,schmidhuber2003godel,zhang2026darwin,wang2026huxley,zhang2026hyperagents}, we must confront a fundamental question: \emph{what is the most foundational and indispensable capability that enables true intelligence?}

We argue that the core of intelligence lies not in the static mastery of a task, but in the \emph{dynamic ability to optimize both the task solution and the optimizer itself}. Consider how humans undertake a sophisticated project: we do not act as a static agent that merely provides solutions. Instead, we engage in a dual-layered refinement, constantly improving the \emph{task solution} we are building while simultaneously sharpening our capability in \emph{optimization}. An engineer does not just write code; they refine their mental models and debugging strategies as the project grows in complexity \citep{flavell1979metacognition,schmidhuber1987evolutionary}. Similarly, a successful organization does not just focus on its immediate output; it evolves its internal structure and decision-making logic to function more effectively \citep{argyris1977double,argyris1978organizational}. If we view the human intellect as an optimizer, this is indeed an optimizer that can optimize itself, i.e. self-referential optimization \citep{good1966speculations, schmidhuber1987evolutionary}. Such self-referential optimization is enabled by the mutual evolution of task solution and optimizer. On the one hand, the optimizer drives the evolution of solutions, with evolved optimization accelerating the evolution. On the other hand, the optimization of optimizer does not occur in a vacuum: it requires the grounding of concrete tasks to provide the empirical feedback necessary to guide the evolution of the optimizer. This mutually beneficial evolution of both the task solution and the optimizer, driven by grounded task execution, is precisely what allows intelligence to transcend the inherent limitations of its initial state.

This dynamic mutual evolution is elegantly captured by the Escherian metaphor of ``Drawing Hands'' \citep{escher1948drawing,hofstadter1979godel}. In Escher’s famous lithograph, two hands emerge from the flat page to draw one another into existence, where each hand is simultaneously the creator and the creation. Building on this philosophy, we propose \textit{Escher-Loop}, a general agent framework that operationalizes this mutual evolution structure. Escher-Loop maintains a self-referential optimization between two distinct populations: (1) a \emph{Task Agent Population} which directly solves a concrete task, e.g., a population of search algorithms, mathematical proofs, or AI customer service agents, and (2) an \emph{Optimizer Agent Population} which aims to refine task agents. The self-referential nature of our framework lies in that the optimization agent population is recursively applied to optimize itself. In this framework, task-level feedback, particularly the empirical performance scores, serves a dual purpose: it guides the refinement of task agents and simultaneously provides the metrics to evaluate and evolve the optimization agents. This closed-loop architecture ensures the system is constantly sharpened in the crucible of task-level execution. By iteratively optimizing the task and optimizer agents through this grounded feedback loop, Escher-Loop minimizes the reliance on human-designed priors and achieves higher performance ceilings and superior adaptability than systems relying on fixed optimization logic.

Our contributions are summarized as follows:

\begin{itemize}
    \item \textbf{The Escher-Loop Framework.} We formalize the foundational capability of an intelligent agent as the recursive optimization of both the task solution and its own optimization capability, grounded by concrete task feedback. Starting from this perspective, we introduce Escher-Loop, a self-referential optimization framework that enables reciprocity between task agents and optimization agents in a fully closed-loop manner. 
    \item \textbf{Dynamic benchmarking without overhead.} To maintain evaluation of optimizers during the continuous evolution of task agents, we propose a dynamic benchmarking mechanism that seamlessly reuses the empirical scores of newly generated task agents as relative win-loss signals to update optimizers’ scores. This mechanism leverages the evolution of task agents as an inherent signal to drive the evaluation and refinement of optimizers without additional overhead.
    \item \textbf{Empirical Validation.} Through experiments, we demonstrate that Escher-Loop autonomously discovers sophisticated optimization strategies that outperform static evolutionary baselines, achieving higher performance ceilings. The code is open-sourced at \url{https://github.com/scaling-group/escher-loop}.
\end{itemize}

\section{Escher-Loop: Closed-Loop Self-Referential Optimization}
\label{sec:methodology}

\subsection{Overview of the Escher-Loop Framework} 

\begin{figure*}
    \centering
    \includegraphics[width=0.7\linewidth]{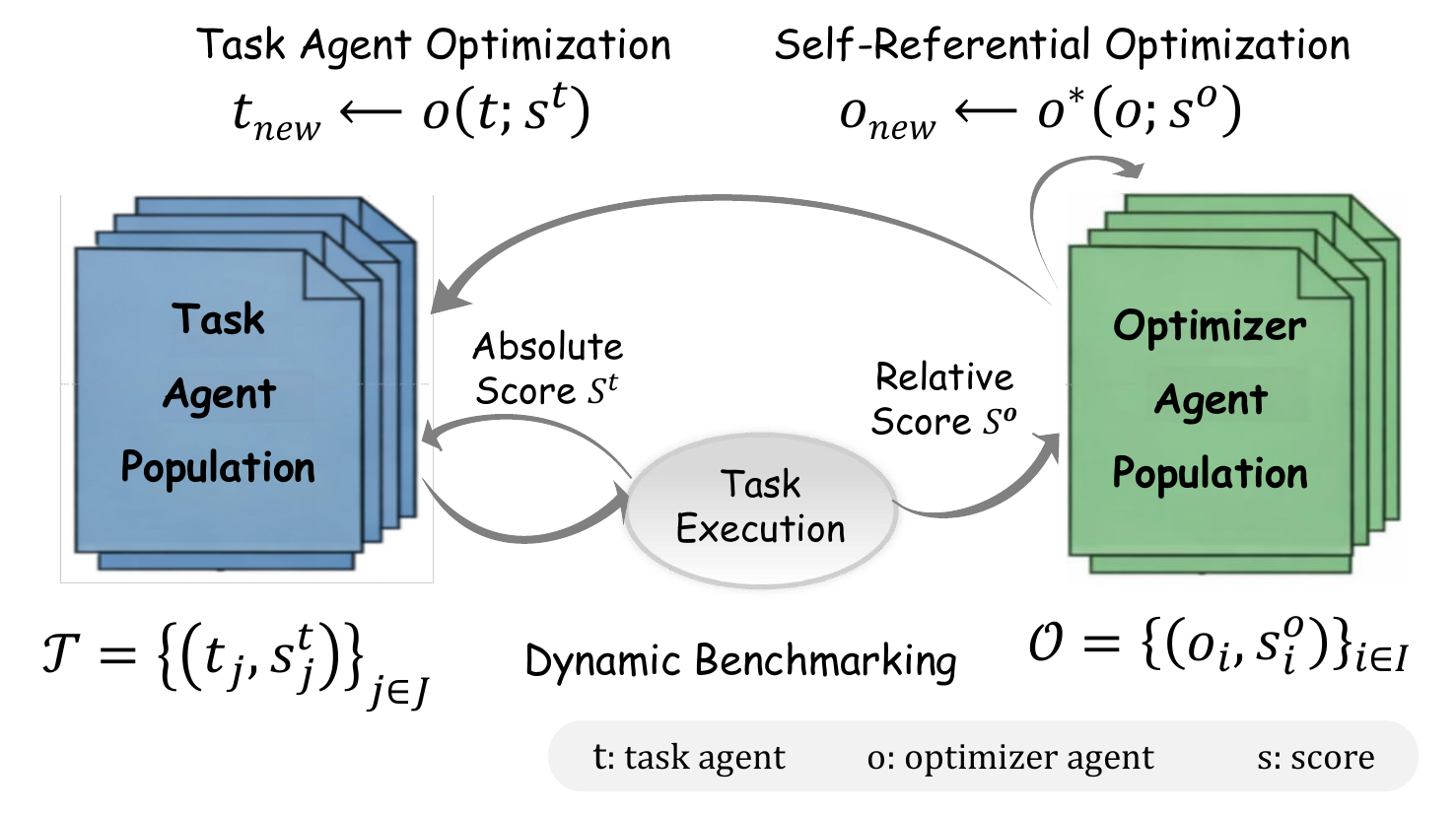}
    \caption{\textbf{Sketch of Escher-Loop.} The Escher-Loop maintains a closed loop between task agents and optimizer agents. When sampled optimizers generate new task-solving agents, their execution outcomes are evaluated to obtain absolute task scores $s^t$. We seamlessly reuse these empirical scores as relative win-loss signals to update the optimizers' scores $s^o$. This dynamic benchmarking mechanism leverages the evolution of task agents as an inherent signal to drive the evaluation and refinement of optimizers without additional overhead, ultimately yielding the joint evolution of both task performance and optimization capability.}

    \label{fig:escher_framework}
\end{figure*}

To transcend the limitations of human-designed priors and static workflows, we propose \textit{Escher-Loop}, a general framework that operationalizes the mutual evolution of task solutions and optimization capabilities. Escher-Loop maintains a self-referential optimization loop between two distinct populations: a Task Agent Population ($\mathcal{T}$) and an Optimizer Agent Population ($\mathcal{O}$). 

Abstractly, within this general framework, a {Task Agent} is a piece of executable code designed to directly solve downstream tasks, and an {Optimizer Agent} is essentially another piece of code designed to optimize code. In the specific implementation presented in this paper based on OpenEvolve \citep{sharma2025openevolve}, we instantiate the optimizer agent as a textual prompt that instructs a Large Language Model (LLM) to perform the optimization, but we highlight that the Escher-Loop framework is general and can be applied to other optimized forms. The framework is initialized with a task evaluation function $f$, along with initial populations $\mathcal{O}_0$ and $\mathcal{T}_0$. By iteratively running a closed-loop execution, the optimizer drives the evolution of task solutions, while the empirical feedback from grounded task execution simultaneously guides the recursive refinement of the optimizers themselves. Ultimately, the algorithm returns the fully optimized populations $\mathcal{O}$ and $\mathcal{T}$ upon convergence.

\subsection{Formalizing the Optimization Step} 
In our algorithm, both populations are explicitly maintained as sets of tuples containing the agents and their respective empirical scores, denoted as $\mathcal{T} = \{(t_j, s^t_j)\}$ and $\mathcal{O} = \{(o_i, s^o_i)\}$. During the evolutionary process, an optimizer agent $o$ is sampled to transform a selected group of task agents into a new, improved candidate. Formally, we represent this generation process by passing the task agents and their corresponding scores together as tuples: 
\begin{equation} 
t_{\text{new}} = o((t_1, s^t_1), \dots, (t_n, s^t_n)) 
\end{equation} 
By explicitly providing the task agents alongside their empirical scores, the optimizer can identify successful optimization strategies and steer the generation toward meaningful improvements. 

Crucially, because an optimizer agent is essentially a piece of code (instantiated as a program that constructs prompts in our implementation), it is fundamentally no different in data modality from the code-based task agents. This structural equivalence means that the optimizer itself can naturally serve as a target for optimization. Just as an optimizer can refine task code, it can seamlessly rewrite and optimize another optimizer's code (or prompt). This property is exactly what enables the self-referential nature of our framework, allowing the optimization agent population to be recursively applied to optimize itself. This self-referential optimization step is formulated as: 
\begin{equation} 
o_{\text{new}} = o((o_1, s^o_1), \dots, (o_m, s^o_m)) 
\end{equation}

\subsection{Dynamic Benchmarking for Optimizers without Overhead}~\label{sec:Dynamic-Benchmarking}
In our framework, evaluating a task agent and maintaining its empirical score $s^t_j$ is straightforward. Because task agents are designed to directly solve concrete problems, their performance can be objectively and easily measured using the provided static task evaluation function $f$. However, accurately evaluating and updating the score $s^o_i$ for each optimizer agent presents a critical challenge. 

Driven by the continuous refinement loop, the task agents are constantly evolving and improving. If we were to evaluate every single optimizer on the most current task agents at every step, the computational cost would be prohibitively high. Indeed, we can only afford to evaluate a subset of optimizer agents at each step. However, this selective evaluation introduces a severe incomparability problem. Because the task agents are constantly being updated, some optimizer agents hold scores derived from optimizing early, naive task agents, while others are evaluated on the most recent, highly refined task agents. The absolute performance improvements achieved on these vastly different generations of task agents cannot be directly compared.

To resolve this discrepancy, we introduce a dynamic benchmarking approach that abandons absolute scoring in favor of relative win-loss outcomes. Instead of measuring absolute score improvements, we apply the sampled subset of optimizers to the same task agents and compare the performance of their newly generated task agents, obtaining pair-wise win-loss signals between the competing optimizers. These win-loss outcomes are integrated into an Elo rating \citep{elo1978rating} system to maintain a globally consistent and fair evaluation across the entire optimizer population over time.

\begin{algorithm}[t]
\caption{Escher-Loop}
\label{alg:escher-loop}
\DontPrintSemicolon
\SetAlgoLined
\linespread{1.1}\selectfont

\KwIn{$f$: task evaluation function; \\
\hspace{1cm} $\mathcal{O}_0, \mathcal{T}_0$: initial populations of (optimizer/task agent, score).}
\KwOut{$\mathcal{O}, \mathcal{T}$: optimized populations of (optimizer/task agent, score).}
\BlankLine

$\mathcal{O} \gets \mathcal{O}_0; \mathcal{T} \gets \mathcal{T}_0$\;

\While{not converged}{
    $\{(o_i, s^o_i)\}_{i \in I} \gets \text{Sample}(\mathcal{O})$; $\{(t_j, s^t_j)\}_{j \in J} \gets \text{Sample}(\mathcal{T})$\; 
    
    \Comment{Optimize and score task agents}
    \For{$i \in I$}{
        $\hat{t}_i \gets o_i(\{(t_j, s^t_j)\}_{j \in J})$\; 
        $\hat{s}^{t}_{i} \gets f(\hat{t}_i)$\;
        $\mathcal{T} \gets \mathcal{T} \cup \{(\hat{t}_i, \hat{s}^{t}_{i})\}$\;
    }
    
    \Comment{Update optimizer scores based on task feedback}
    $W \in \{-1, 0, 1\}^{|I| \times |I|}\gets \text{PairwiseCompetition}(\{\hat{s}^t_i\}_{i \in I})$\; 
    $\{s^o_i\}_{i \in I} \gets \text{EloUpdate}( W, \{s^o_i\}_{i \in I})$\; 
    \BlankLine
    \Comment{Self-referential optimization}
    $(o^*, {-}), \{(o_i, s^o_i)\}_{i \in \hat{I}} \gets \text{Sample}(\mathcal{O})$\; 
    $\hat{o} \gets o^*(\{(o_i, s^o_i)\}_{i \in \hat{I}})$\; 
    $\hat{s}^o \gets \text{Initialize}(\{s^o_i\}_{i \in \hat{I}})$\;
    $\mathcal{O} \gets \mathcal{O} \cup \{(\hat{o}, \hat{s}^o)\}$\;
}
\Return{$\mathcal{O}, \mathcal{T}$}
\end{algorithm}

Notably, our Elo-based dynamic benchmarking not only allows the system to robustly maintain optimizer scores despite the shifting evaluation landscape, but also substantially improves resource allocation without computational overhead. Instead of drawing subsets uniformly at random, we predominantly sample top optimizer agents with the explicit goal of generating better task agents. As a mandatory step, the generated new task agents are evaluated by $f$ to obtain their empirical scores $s^t$. We seamlessly reuse these exact same scores to determine relative win-loss outcomes and update the participating optimizers' Elo ratings $s^o$, establishing a more fine-grained ranking specifically among the sampled top-performing optimizers. This elegant dual-use mechanism ensures that we focus on driving the task agent evolution with the most promising optimization agents, while also evaluating and differentiating these optimizers almost ``for free''.

\subsection{The Closed-Loop Mutual Evolution}
To synthesize the mechanisms above, Escher-Loop operates as a continuous mutual evolution loop. The process is initialized with optimizer and task populations, $\mathcal{O}_0$ and $\mathcal{T}_0$, together with the task evaluation function $f$, and then alternates between improving task agents and improving the optimizer population that produces them.

At each iteration, the algorithm samples subsets from both populations using a unified rank-based Softmax mechanism. By relying on relative ranks rather than absolute score magnitudes, the sampling remains robust under heterogeneous and shifting score scales, while the temperature parameter controls the exploration--exploitation trade-off.

The sampled optimizers generate new task candidates from the sampled task agents. These candidates are immediately evaluated by $f$ and integrated into $\mathcal{T}$. The same empirical task scores then serve a second role: they provide win-loss evidence for updating the participating optimizers' Elo ratings through the dynamic benchmarking procedure, thereby linking optimizer selection to grounded task feedback rather than to a separate evaluator.

The loop then executes the self-referential step. A lead optimizer $o^*$ rewrites a sampled subset of the optimizer population to synthesize a new optimizer candidate $\hat{o}$, whose initial score is inherited from its predecessors before it is added back into $\mathcal{O}$. This final step makes the optimizer population itself an evolving object rather than a fixed search heuristic.

This cycle of rank-based sampling, task generation, dual-purpose evaluation, and self-referential optimizer evolution continuously sharpens both concrete task agents and optimization capabilities. The overall procedure is summarized in Algorithm \ref{alg:escher-loop}.

\section{Results and Analysis}
\label{sec:experiments}

\begin{figure*}[h]
    \centering
    \includegraphics[width=\linewidth]{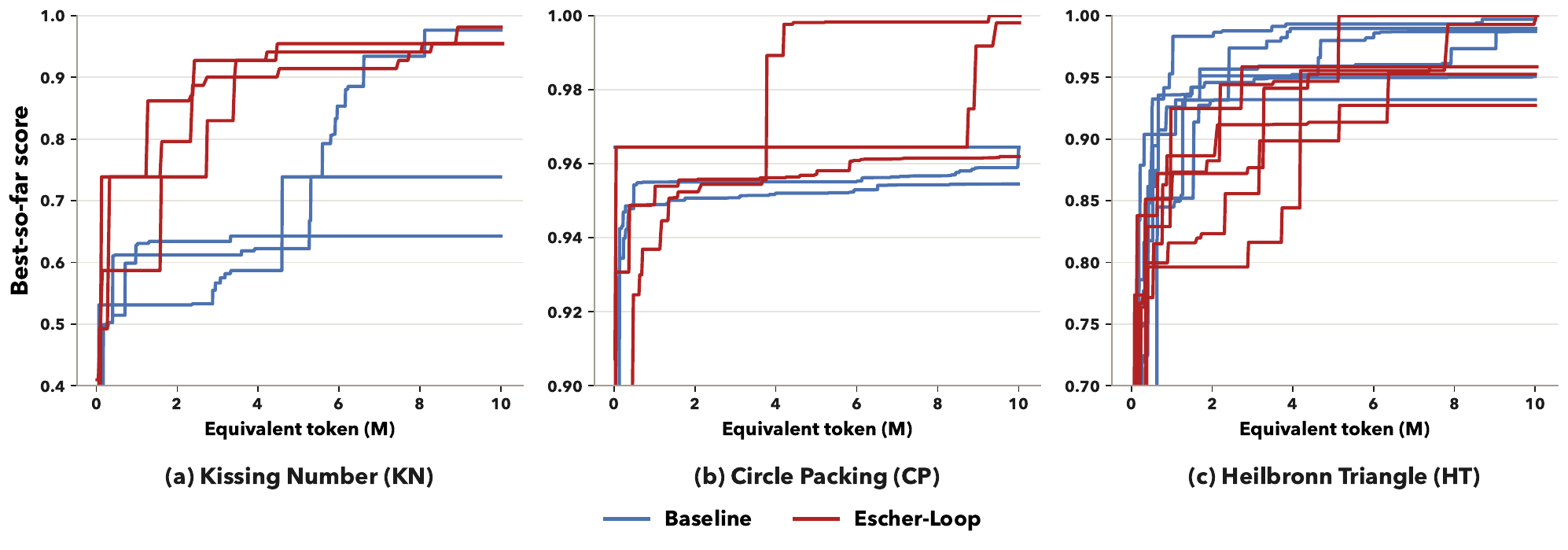}
    \caption{Comparison of best-so-far task performance between the handcrafted baseline optimizer and Escher-Loop across three optimization landscapes: Kissing Number (KN), Circle Packing (CP), and Heilbronn Triangle (HT). Each curve represents one independent run, and the reported score is the best-so-far normalized task score as a function of cumulative equivalent token consumption.}
    \label{fig:main_baseline_vs_escher}
\end{figure*}

\begin{figure*}[h]
    \centering
    \includegraphics[width=\linewidth]{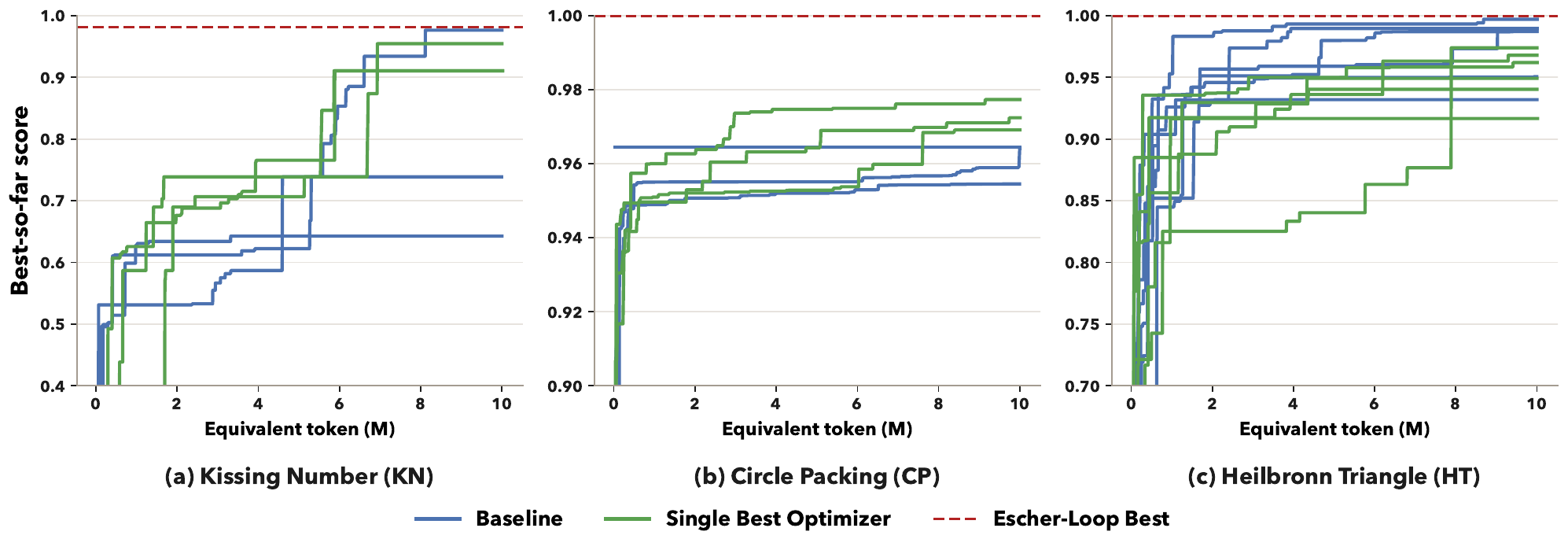}
    \caption{Comparison of best-so-far normalized task performance between the handcrafted baseline optimizer and the single best evolved optimizer across three optimization landscapes: Kissing Number (KN), Circle Packing (CP), and Heilbronn Triangle (HT). Each curve represents one independent run, while the horizontal dashed line indicates the best performance achieved by Escher-Loop.}
    \label{fig:main_baseline_vs_single_best}
\end{figure*}

We evaluate Escher-Loop along two complementary axes: (1) whether closed-loop self-referential optimization improves task-level search under a controlled compute budget compared to static baselines, and (2) how the underlying mechanisms, specifically the prioritization driven by dynamic benchmarking and the autonomous emergence of sophisticated optimization strategies, enable this superior performance.

\paragraph{Experimental Protocol} We consider three geometric optimization landscapes: Kissing Number (KN), Circle Packing (CP), and Heilbronn Triangle (HT). To ensure a fair cost comparison, all methods are evaluated under a strict budget of 10M equivalent tokens, a metric that normalizes the combined cost of input and output tokens into a single output-token equivalent based on relative API pricing. Performance is reported using normalized task scores for the Task Agent Population and Elo ratings for the Optimizer Agent Population. Detailed experimental setups can be found in Appendix~\ref{sec:appendix_setup}.

\subsection{Task Performance}
\label{subsec:main_results}

We employ OpenEvolve~\citep{sharma2025openevolve}, an open-source reproduction of AlphaEvolve, as the representative handcrafted baseline optimizer. Escher-Loop is compared against this static baseline across the three optimization landscapes. To ensure a fair comparison of computational cost, the evolving trajectories are evaluated by plotting the best-so-far normalized task score as a function of token consumption on the horizontal axis.

\paragraph{Results} As illustrated in Figure~\ref{fig:main_baseline_vs_escher}, Escher-Loop consistently improves the optimization ceiling on the harder KN and CP landscapes, where the handcrafted baseline often enters long plateaus. On HT, the baseline reaches high scores quickly, but Escher-Loop still attains the highest peak score among our runs.

\paragraph{Analysis} These results suggest that fixed handcrafted heuristics can be efficient on simpler landscapes, but are less robust on tasks where the useful search strategy changes over time. Escher-Loop benefits from evolving the optimization process itself, allowing it to escape suboptimal regimes and reach higher task scores.

\subsection{The Necessity of Evolving Optimizer Population} 
\label{subsec:dynamic_vs_static}

To evaluate the intrinsic capability of the evolved optimizers, we isolate the highest-scoring agent from the Escher-Loop evolution process, termed the \emph{Single Best Optimizer (SBO)}. By substituting the handcrafted optimizer in OpenEvolve with this SBO, we directly compare their isolated trajectories. Note that the best performance developed by Escher-Loop is denoted by a horizontal dashed upper bound.

\paragraph{Results} As shown in Figure~\ref{fig:main_baseline_vs_single_best}, the SBO transfers useful search behavior on KN and CP, often outperforming the handcrafted baseline. However, it underperforms the baseline on HT, and across all three landscapes it never exceeds the peak score achieved by the full Escher-Loop framework.

\paragraph{Analysis} The SBO result shows that self-referential evolution can produce strong optimizer agents, but also that a single frozen optimizer does not reproduce the full system. The advantage of Escher-Loop comes from maintaining an evolving optimizer population whose search policy can keep adapting as the task population changes.

\subsection{Mechanism Ablations}
\label{subsec:mechanism_ablations}

The previous results show that Escher-Loop improves over both the handcrafted baseline and a single transferred optimizer. We next test which parts of the closed-loop mechanism are responsible for this improvement. Since Escher-Loop exhibits the clearest gains on Kissing Number (KN) and Circle Packing (CP), we conduct mechanism ablations on these two tasks.

We consider three ablated variants. \emph{\textbf{Random Elo Update}} preserves the rating interface but breaks its semantic connection to task feedback: after task execution, optimizer ratings are randomly perturbed rather than updated according to the observed relative task outcomes. This approximates an optimizer population whose selection pressure is effectively random. \emph{\textbf{Fixed Elo = 1200}} removes relative optimizer differentiation by keeping all optimizer ratings fixed, so optimizer agents are no longer ranked by empirical evidence from task improvement. \emph{\textbf{Static-50 Optimizer Pool}} preserves optimizer diversity but removes self-referential optimizer evolution: we initialize a fixed pool of 50 manually selected optimizer agents, matching the optimizer population cap used in Escher-Loop, and use this pool to optimize task agents without evolving the optimizer population itself. This ablation tests whether population diversity alone explains the gains.

Each ablation is repeated three times with the same 10M equivalent-token budget used in the main Baseline and Escher-Loop comparisons. We summarize performance using two metrics, reporting the mean across runs. \emph{Best@10M} is the best normalized task score reached within the budget. \emph{AUC@10M} is the area under the best-so-far task-score curve over the same budget, normalized by the budget:
\begin{equation}
\mathrm{AUC@10M}=\frac{1}{10M}\int_{0}^{10M} \mathrm{best\_so\_far}(t)\,dt .
\end{equation}
In practice, this integral is computed from the recorded discrete trajectory as a step-function area. Thus, Best@10M measures the final achieved solution quality, while AUC@10M measures search efficiency throughout the run. Since several tasks exhibit discrete performance plateaus, run-to-run variability in final best score can be visually misleading; we therefore report variability through AUC standard deviation, which captures differences in when high-quality solutions are discovered.

\begin{table}[h]
\caption{Mechanism ablations on Kissing Number and Circle Packing. Each ablation is repeated three times. Best and AUC report mean values under the 10M-token budget, following the same repeated-run protocol as the main comparisons; AUC Std summarizes run-to-run stability.}
\label{tab:mechanism_ablations}
\centering
\begin{subtable}[t]{0.49\linewidth}
\centering
\caption{Kissing Number (KN)}
\label{tab:ablation_kn}
\resizebox{\linewidth}{!}{
\begin{tabular}{lccc}
\toprule
Method & Best $\uparrow$ & AUC $\uparrow$ & AUC Std $\downarrow$ \\
\midrule
Baseline & 0.786 & 0.674 & 0.067 \\
Single Best Optimizer & 0.925 & 0.738 & 0.048 \\
Static-50 Optimizer Pool & 0.568 & 0.408 & 0.181 \\
Fixed Elo = 1200 & 0.811 & 0.772 & 0.074 \\
Random Elo Update & 0.739 & 0.721 & 0.016 \\
\midrule
Escher-Loop (ours) & \textbf{0.963} & \textbf{0.876} & 0.032 \\
\bottomrule
\end{tabular}
}
\end{subtable}
\hfill
\begin{subtable}[t]{0.49\linewidth}
\centering
\caption{Circle Packing (CP)}
\label{tab:ablation_cp}
\resizebox{\linewidth}{!}{
\begin{tabular}{lccc}
\toprule
Method & Best $\uparrow$ & AUC $\uparrow$ & AUC Std $\downarrow$ \\
\midrule
Baseline & 0.961 & 0.956 & 0.008 \\
Single Best Optimizer & 0.973 & 0.962 & 0.007 \\
Static-50 Optimizer Pool & \textbf{0.988} & \textbf{0.966} & 0.009 \\
Fixed Elo = 1200 & 0.969 & 0.959 & 0.003 \\
Random Elo Update & 0.964 & 0.961 & 0.001 \\
\midrule
Escher-Loop (ours) & 0.987 & \textbf{0.966} & 0.009 \\
\bottomrule
\end{tabular}
}
\end{subtable}
\end{table}

\paragraph{Analysis} The ablations indicate that Escher-Loop's gains do not arise merely from using more optimizer agents. On KN, the Static-50 pool has the largest variance and substantially lower AUC, showing that a fixed diverse optimizer pool can occasionally discover useful search directions but fails to provide stable optimization pressure. Fixed Elo and Random Elo preserve the appearance of an optimizer population, yet both remove the grounded relative selection signal that connects optimizer quality to task feedback; they reach the common KN plateau but fall below Escher-Loop in both final performance and search efficiency. On CP, the baseline is already strong and all methods are closer in absolute score; Static-50 attains a slightly higher mean peak score, while Escher-Loop remains effectively tied in AUC and substantially stronger on KN. Together, these results support the central claim that optimizer diversity alone is insufficient: it must be coupled with grounded, dynamic selection and self-referential optimizer evolution.

\subsection{Qualitative Analysis of Emergent Optimization Strategies}
\label{subsec:qualitative_analysis}

\begin{figure*}[p]
    \centering
    \captionsetup[subfigure]{font=normalsize,skip=4pt}
    \setlength{\abovecaptionskip}{6pt}
    \setlength{\belowcaptionskip}{0pt}

    \begin{subfigure}[b]{0.48\textwidth} 
        \centering
        \includegraphics[width=\textwidth]{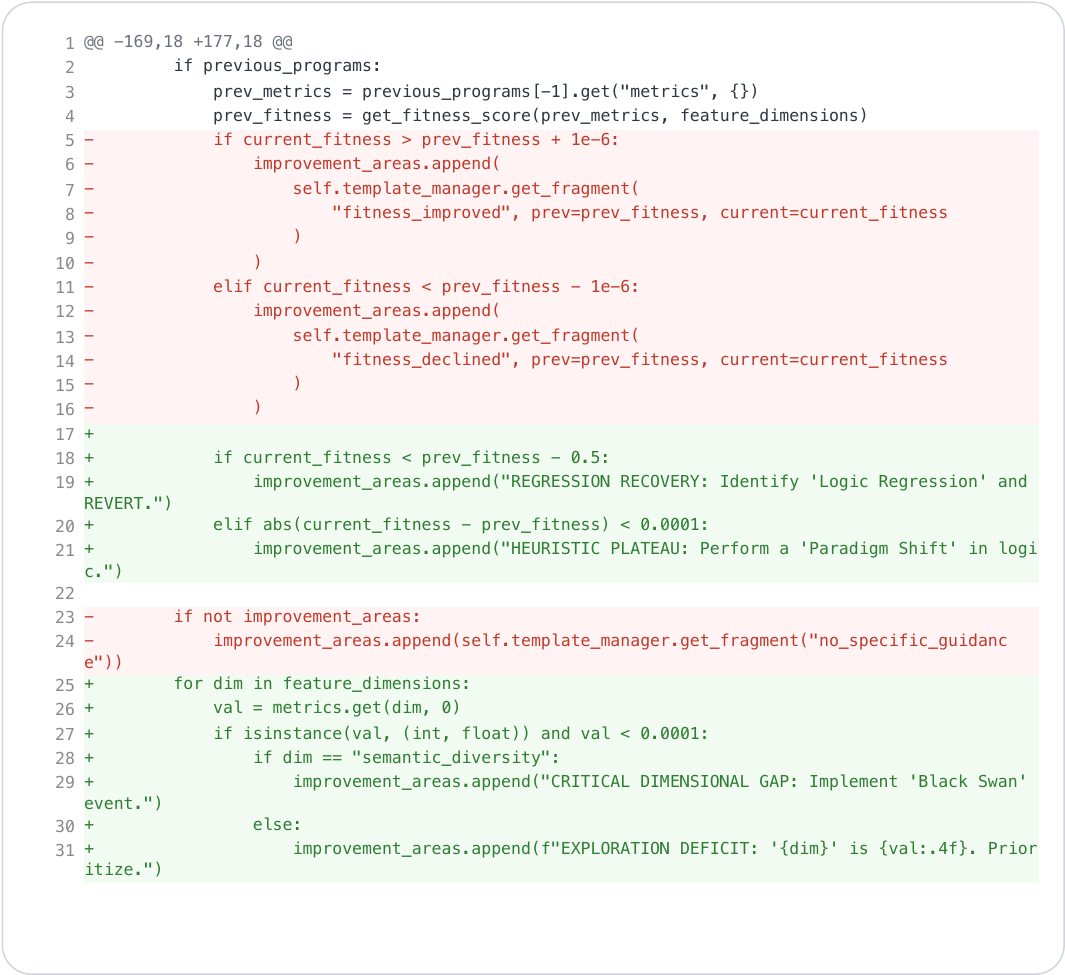}
        \captionsetup{width=0.92\linewidth} 
        \caption{\textbf{Diagnostic feedback mechanism}  The optimizer replaces generic fitness-improvement messages with explicit recovery and plateau signals, such as regression recovery and low-diversity warnings.}
        \label{fig:sub_a}
    \end{subfigure}
    \hfill 
    \begin{subfigure}[b]{0.48\textwidth} 
        \centering
        \includegraphics[width=\textwidth]{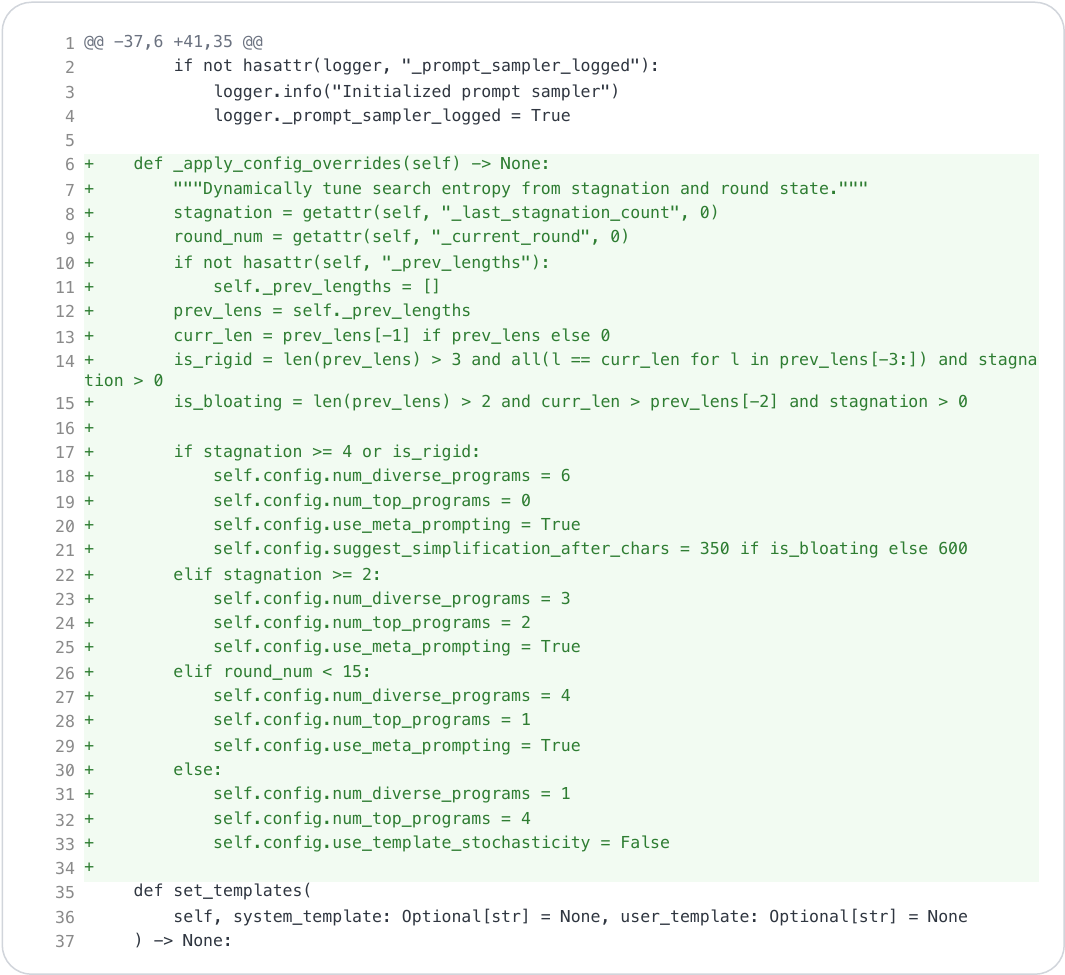}
        \captionsetup{width=0.92\linewidth}
        \caption{\textbf{Adaptive search control}  The optimizer adjusts exploration pressure according to stagnation, code rigidity, bloat, and evolution stage, shifting between high-diversity exploration and elite-focused refinement.}
        \label{fig:sub_b}
    \end{subfigure}
    \vspace{1.2em} 
    \begin{subfigure}[b]{0.48\textwidth} 
        \centering
        \includegraphics[width=\textwidth]{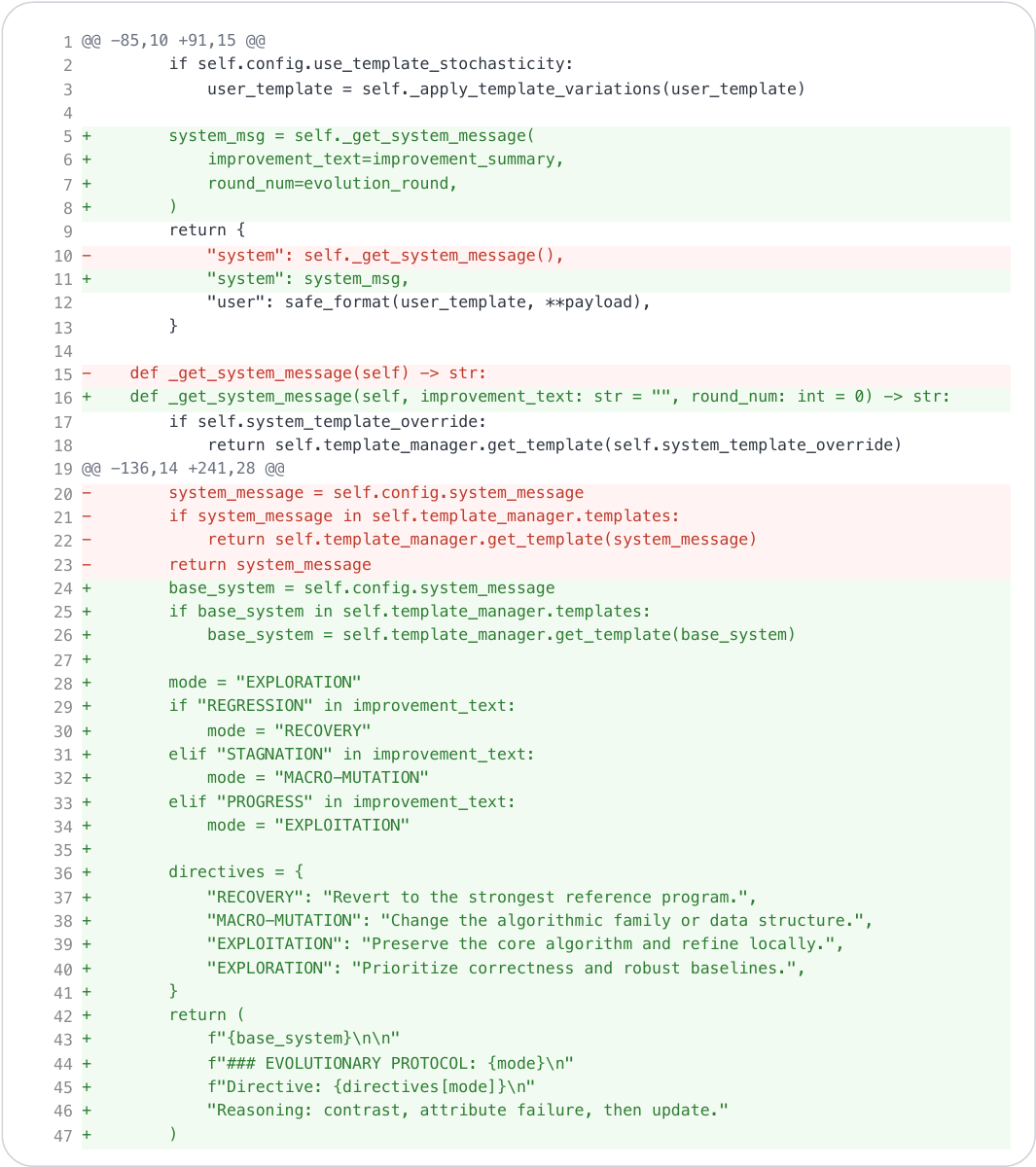}
        \captionsetup{width=0.92\linewidth}
        \caption{\textbf{Stage-aware system prompting}  The optimizer selects different prompt protocols for exploration, regression recovery, macro-mutation, and exploitation, making the system message conditional on recent optimization signals.}
        \label{fig:sub_c}
    \end{subfigure}
    \hfill 
    \begin{subfigure}[b]{0.48\textwidth} 
        \centering
        \includegraphics[width=\textwidth]{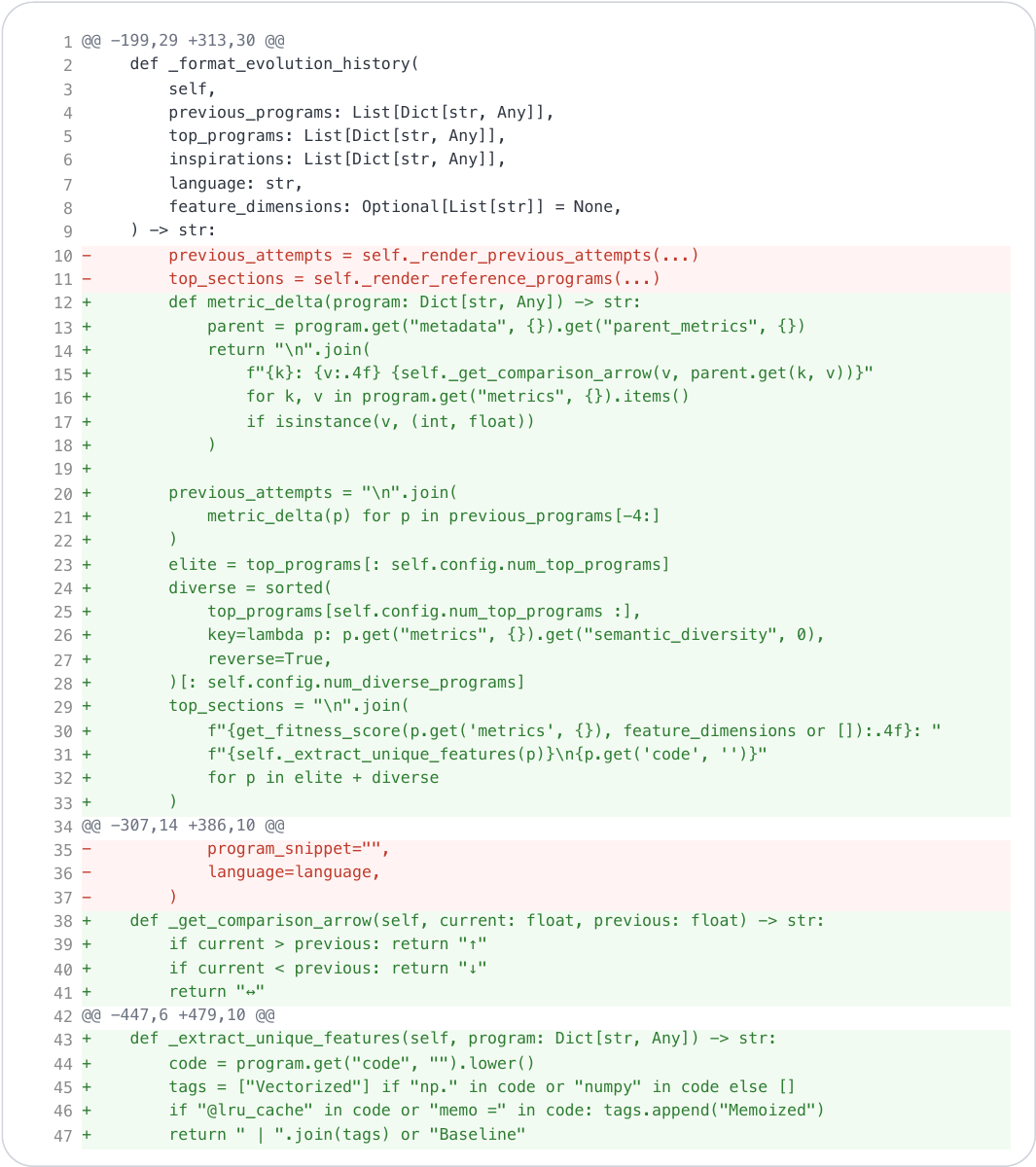}
        \captionsetup{width=0.92\linewidth}
        \caption{\textbf{Reference-program mining}  The optimizer expands evolution history into structured references by tracking metric deltas, selecting elite and diverse programs, and extracting reusable architectural features from prior candidates.}
        \label{fig:sub_d}
    \end{subfigure}
    \caption{\textbf{Representative code segments from evolved optimizer agents.}
    The four examples illustrate how self-referential optimization reshapes the optimizer's prompt-construction logic by introducing diagnostic feedback, adaptive search control, stage-specific prompting, and reference-program mining. Together, these snippets provide qualitative evidence that the optimizer population develops increasingly sophisticated strategies for interpreting task feedback and structuring future search, rather than relying solely on a fixed handcrafted prompt.}
    \label{fig:total_optimizer}
\end{figure*}

The continuous improvements in task performance observed in our experiments emerge from the synergistic mutual evolution between the Task Agent and Optimizer Agent populations. As the optimizers generate stronger task agents, the shifting evaluation landscape in turn drives the optimizers to evolve more capable optimization strategies. As anticipated by our closed-loop design, the optimizers do not merely accumulate code complexity; rather, they develop increasingly sophisticated optimization strategies. As illustrated by the representative code segments in Figure~\ref{fig:total_optimizer}, the evolved optimizer transcends basic LLM prompting. It shifts away from the fixed logic of handcrafted baselines and develops strategies that can process rich evolutionary feedback to make informed search decisions.

We observe that this capacity for autonomous evolution naturally extends across all three problem domains considered in this work. Specifically, they discover intuitive and highly effective search strategies, such as reasoning over historical attempts to avoid repeated failures, and extracting beneficial logic from diverse candidate programs to combine them. This consistent ability to dynamically synthesize and deploy context-appropriate optimization strategies demonstrates the effectiveness of our self-referential framework.

\section{Related Works}\label{sec:related_works}
\paragraph{Language Models as Optimizers} 
Large language models (LLMs) are naturally suited for gradient-free optimization, ranging from optimizing natural language prompts \citep{yang2024large,fernando2024promptbreeder} to evolving agent scaffolds like DAG configurations \citep{hu2025automated,liu2026todoevolve} and optimizing initialization seed \citep{tan2026agentga}. Another prominent direction treats program search as an evolutionary cycle \citep{schmidhuber1987evolutionary}, where frameworks like FunSearch \citep{romera2024mathematical} and AlphaEvolve \citep{novikov2025alphaevolve} use LLMs as code-level mutation operators for algorithmic discovery. Subsequent extensions, whether incorporating heuristic modifications like memory archives, such as OpenEvolve \citep{sharma2025openevolve}, ThetaEvolve \citep{wang2025thetaevolve}, ShinkaEvolve \citep{lange2026shinkaevolve}, EvoX \citep{liu2026evox}, and TTT-Discover \citep{yuksekgonul2026learning}, designing adaptive search controllers, such as AdaEvolve \citep{cemri2026adaevolve}, or scaling these evaluation loops like SimpleTES \citep{ye2026evaluation}, remain entangled with human-engineered heuristics and confined to static workflows.
\paragraph{Recursive Self-Improvement (RSI)} 
While using LLMs as static optimizers yields strong performance, the pursuit of open-ended growth requires systems that can improve their own optimization mechanisms. Pioneered by \citet{good1966speculations} and later formalized by the Gödel Machine \citep{schmidhuber2003godel}, the concept of self-referential systems has been fundamentally explored in neural networks \citep{kirsch2022eliminating}. Recently, this pursuit has transitioned into empirical LLM-driven code-level self-modification. Frameworks such as STOP \citep{zelikman2024selftaught}, Gödel Agent \citep{yin2025godelagent}, and SICA \citep{robeyns2025selfimproving} eliminate the boundary between target and meta-agents, enabling systems to dynamically diagnose and rewrite their own codebases. To overcome the limitations of greedy single-path updates, DGM \citep{zhang2026darwin} introduces a growing tree of historical agent variants, while HGM \citep{wang2026huxley} utilizes clade-based meta-productivity to balance exploration. Further abstracting this capability, Hyperagents \citep{zhang2026hyperagents} unifies task and meta-level logic into an editable program library, enabling cross-domain meta-cognitive optimization.
\paragraph{Population Evolution} 
To overcome the limited exploration and overfitting of isolated individual RSI, recent works such as GEA \citep{weng2026group} and TerraLingua \citep{paolo2026terralingua} expand the evolutionary unit to populations by introducing lateral experience transfer and environmental dynamics. Pushing this toward full autonomy, CORAL \citep{qu2026coral} enables asynchronous agents to co-evolve via shared persistent memory without fixed heuristics. More closely related to our closed-loop approach, frameworks like CoEvolve \citep{yang2026coevolve} and Paso Doble \citep{zhang2026better} introduce \textit{mutual evolution}, where the optimizer continuously synthesizes targeted adversarial tasks to co-evolve alongside the policy.

\section{Conclusion}
\label{sec:conclusion}

In this work, we shift the focus of autonomous agents from task-level solutions to \emph{optimization ability} itself, and introduce {Escher-Loop}, a closed-loop, self-referential framework in which optimization becomes the object of evolution. Our results suggest that optimization ability can be treated as a comparable and continuously improvable property, rather than a fixed mechanism prescribed by human design.

Beyond the empirical results, our findings further illuminate how such improvement becomes possible. The closed-loop coupling between optimizer agents and a dynamically evolving task population creates a continually shifting evaluation landscape, in which optimization strategies are repeatedly tested, compared, and refined under relative feedback. This interaction induces a form of pressure analogous to natural selection, enabling optimization strategies to adapt and accumulate improvements over time without explicit supervision or retraining. 
As a result, self-referential improvement emerges not as a predefined objective, but as a natural outcome of the synergistic mutual evolution within this closed-loop system.

More broadly, Escher-Loop highlights a shift in objective from a single task solution to the iterative optimization process itself. While our current exploration demonstrates self-referential improvement within individual problem spaces, it opens the door to an even more ambitious vision: \emph{cross-domain multi-task optimization}. In such future systems, generalization would no longer be achieved solely through improved task-specific performance, but through a unified, transferable optimization ability that governs how solutions are updated across diverse tasks and environments. 
We believe this perspective opens a path toward more autonomous and scalable forms of intelligence, in which continual improvement emerges as an intrinsic property of the system. We leave the full validation of this unified view and the achievement of true cross-domain transferability to future work.

\section*{Acknowledgments}
Ziyang Liu, Xinyan Guo, and Xuchen Wei were supported by Shenzhen X-Institute through its Pioneering Investigator Program. Liu Yang acknowledges support from the National Research Foundation, Singapore, under the NRF fellowship (Project No. NRF-NRFF17-2025-0006).

\bibliographystyle{unsrtnat}
\bibliography{references}

\newpage
\appendix

\section{Implementation Details and Experimental Protocol}
\label{sec:appendix_setup}

This appendix provides implementation details that support reproducibility and clarify the experimental protocol underlying the main results in Section~\ref{sec:experiments}. Specifically, we summarize the compute normalization used in matched-budget comparisons, the model-routing and generation configurations, the prompt interface exposed to optimizer agents, and the archive and population settings used in closed-loop evolution.

\subsection{Compute Normalization via Equivalent Tokens}
\label{subsec:appendix_equivalent_token}

All principal comparisons are conducted under a matched \emph{equivalent-token} budget. Because, in the API setting used in our experiments, input tokens are priced at one quarter of output tokens, we normalize total token consumption into a unified output-equivalent measure:
\begin{equation}
T_{\mathrm{eq}} = T_{\mathrm{out}} + 0.25\,T_{\mathrm{in}},
\end{equation}
where $T_{\mathrm{in}}$ and $T_{\mathrm{out}}$ denote the consumed input and output tokens, respectively. This normalization ensures that performance differences cannot be attributed merely to unequal expenditure of inference-time compute. Unless otherwise stated, the matched-budget comparisons in the main text are reported under a budget of 10M equivalent tokens per task.

\subsection{Model Ensemble and Generation Settings}
To balance reasoning quality, throughput, and latency, we employ a hybrid Gemini 3 Flash \citep{deepmind2025gemini3flash} ensemble. The generation and routing configuration is summarized below.

\begin{itemize}
    \item \textbf{Model allocation.} We route 80\% of requests to \texttt{gemini-3-flash-preview} with \texttt{thinkingLevel} set to \texttt{low}, while the remaining 20\% use the model's default dynamic thinking configuration. This asymmetric allocation provides a practical trade-off between throughput, cost, and reasoning depth.

    \item \textbf{Temperature control.} We use a two-level temperature scheme.
    \begin{enumerate}
        \item \textit{Generation temperature.} During optimizer evolution, we use a generation temperature of $T=1.0$ to encourage exploratory variation in candidate proposals. During downstream task optimization, we reduce this temperature to $T=0.7$ to improve stability and reduce unproductive variance.
        \item \textit{Archive sampling temperature.} The MAP-Elites archive \citep{mouret2015illuminating} uses rank-based Softmax sampling, $P(i) \propto \exp(-\mathrm{rank}_i/T)$, with role-specific temperatures:
        \begin{itemize}
            \item \textbf{Matchmaking ($T=1.2$):} encourages moderate exploration when selecting opponents of comparable Elo level;
            \item \textbf{Mentoring ($T=0.5$):} emphasizes stronger optimizer agents as teachers, thereby reducing entropy during guidance;
            \item \textbf{Checkpoint selection ($T=1.2$):} balances exploitation of high-quality archive states against exploration of historically diverse ones.
        \end{itemize}
    \end{enumerate}
    
    \item \textbf{Sequence length.} The \texttt{max\_tokens} limit is fixed at 60{,}000 across all phases.
    
    \item \textbf{Request management.} We use a timeout of 1{,}200\,s during optimizer evolution and 600--800\,s during downstream task optimization, with at most three retries per request.
\end{itemize}

\subsection{Rank-Based Population Sampling}
Whenever Escher-Loop samples from either the task-agent or optimizer-agent population, it uses rank-based Softmax sampling. Given a candidate set $\mathcal{C}$, each element $i \in \mathcal{C}$ is assigned a rank $\mathrm{rank}(i)$ according to its score, where better candidates receive smaller ranks. The sampling probability is
\begin{equation}
P(i)=
\frac{\exp\left(-\mathrm{rank}(i)/\tau\right)}
{\sum_{j \in \mathcal{C}} \exp\left(-\mathrm{rank}(j)/\tau\right)},
\end{equation}
where $\tau$ controls the exploration--exploitation trade-off. This rank-based form avoids relying on absolute score magnitudes, which may be incomparable across evolving task-agent and optimizer-agent populations.



\subsection{Optimizer Evolution and MAP-Elites Configuration}
The closed-loop system maintains an archive-based optimizer population with the following settings:

\begin{itemize}
    \item \textbf{Population control.} The optimizer population size is set to 50, and checkpoints are saved every 20 iterations.
    \item \textbf{Elo rating.} Optimizer agents are initialized with Elo 1200.0 and updated using a $K$-factor of 32.0.
    \item \textbf{Archive usage.} The optimizer population is maintained in a MAP-Elites-style archive, where both fitness and behavioral diversity are preserved. This archive is used not only for selection, but also as contextual input to subsequent optimization steps through elite references, diverse inspirations, and historical checkpoints.
    \item \textbf{Dynamic benchmarking.} Optimizer quality is assessed through repeated relative comparisons under evolving task populations rather than through a single static benchmark. Elo is therefore used as a comparative ranking signal for optimizer agents, while downstream task metrics remain the ground-truth signal for task-program quality.
\end{itemize}

\section{Dynamic Benchmarking Diagnostics}
\label{sec:appendix_dynamic_benchmarking}

This appendix provides an auxiliary diagnostic view of the dynamic benchmarking mechanism described in Section~\ref{sec:Dynamic-Benchmarking}. The main paper evaluates the importance of this mechanism through the ablation results in Section~\ref{subsec:mechanism_ablations}. Here, we additionally visualize the optimizer-rating dynamics that arise during closed-loop evolution.

\begin{figure*}[h]
    \centering
    \includegraphics[width=\linewidth]{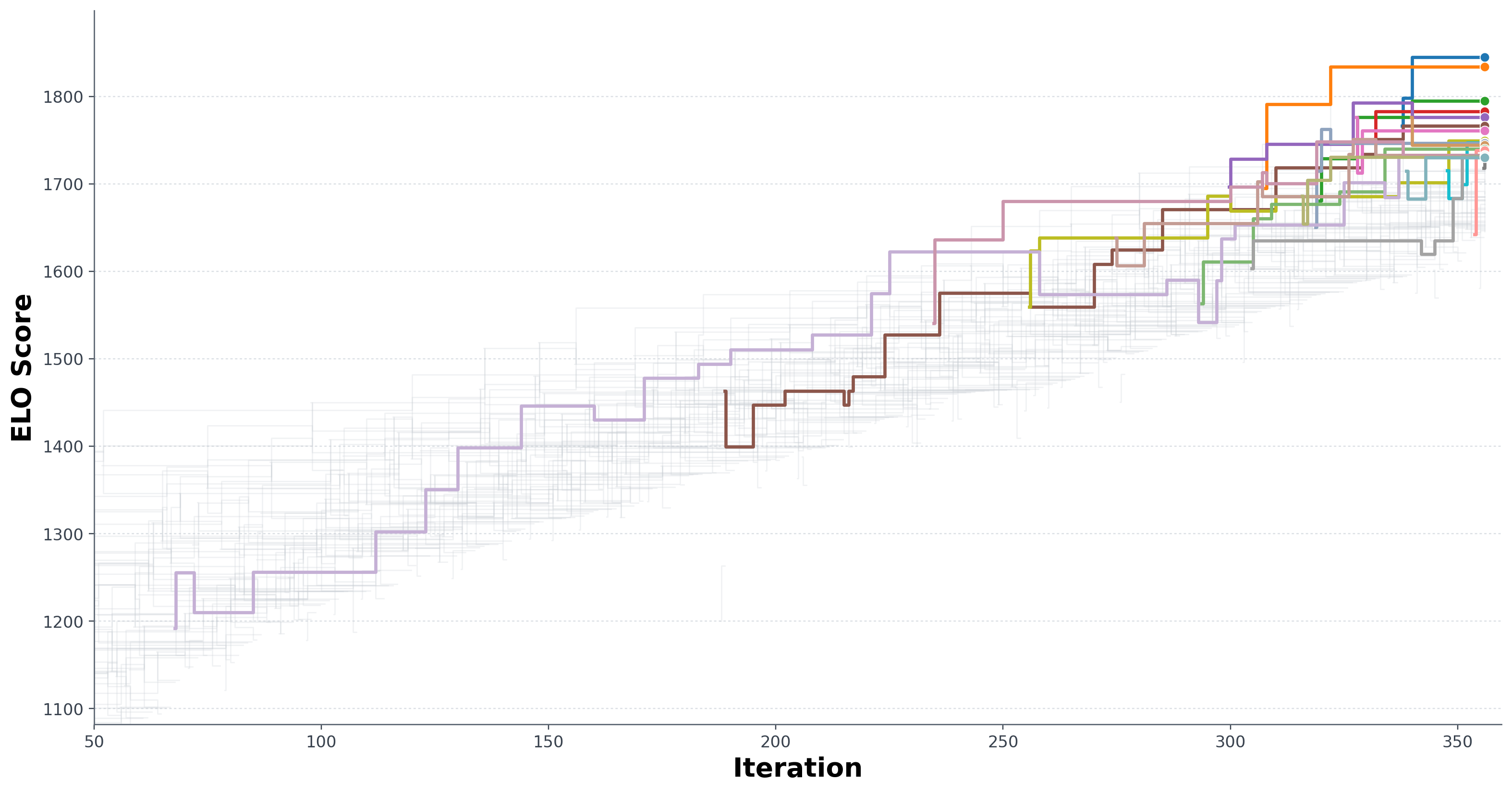}
    \caption{\textbf{Optimizer Elo trajectories during closed-loop evolution.}
    Colored lines denote the 20 optimizer agents with the highest final Elo, while gray lines show the remaining optimizer trajectories. The active optimizer population is capped at 50 agents: new agents are spawned during evolution, and weaker agents can be removed from the active population. The stepwise pattern occurs because Elo is updated only when an optimizer is sampled for relative evaluation. Top optimizers are sampled more often, receive denser evidence, and are more likely to remain active.}
    \label{fig:elo_traj_appendix}
\end{figure*}

Because the task-agent population continuously improves, absolute task scores are not a stationary measure of optimizer quality. Dynamic benchmarking therefore compares optimizers through relative outcomes on the same sampled task context, and uses the resulting win-loss evidence to update optimizer ratings. The trajectories in Figure~\ref{fig:elo_traj_appendix} are not used as a standalone proof that Elo is a perfect measure of optimization ability. Instead, they illustrate the operational role of Elo in Escher-Loop: it provides a continually updated selection signal that allocates more task-generation opportunities to optimizers with stronger recent empirical evidence while still permitting lower-ranked optimizers to re-enter through stochastic sampling.
\section{Best Evolved Task Programs}
\label{sec:appendix_best_programs}

This appendix presents the best-performing task agents discovered by Escher-Loop on the three optimization landscapes studied in the paper. For each task, we first restate the formal problem definition, including its search domain, task-specific parameters, and evaluation rule, and then show the corresponding program. Throughout this appendix, the normalized score is computed as
\begin{equation}
s_{\mathrm{norm}}=\frac{s_{\mathrm{raw}}}{s_{\mathrm{ref}}},
\end{equation}
where $s_{\mathrm{raw}}$ is the task-native evaluator output and $s_{\mathrm{ref}}$ is the reference value used for normalization in the main comparisons.
These reference values $s_{ref}$ are fixed benchmark constants adopted directly from the AlphaEvolve evaluation environments \citep{novikov2025alphaevolve} to ensure standardized comparisons. Note that $s_{ref}$ represents a truncated baseline; consequently, the unclipped normalized score $s_{norm}$ may slightly exceed 1.0 due to this precision difference.
\subsection{Kissing Number (KN)}
For Kissing Number, the search space consists of finite sets of nonzero integer vectors in $\mathbb{Z}^{11}$. A candidate set $S$ is valid only if every point satisfies the norm bound
\begin{equation}
\|x\|_2^2 \le 4, \qquad x \in S,
\end{equation}
and every pair of distinct points satisfies the separation constraint
\begin{equation}
\|x-y\|_2^2 \ge 4, \qquad x \neq y,\; x,y \in S.
\end{equation}
The evaluator returns the cardinality of the valid construction,
\begin{equation}
s_{\mathrm{raw}} = |S|.
\end{equation}
In our experiments, the task parameter is fixed to dimension $d=11$, and the reference value used for normalization is $s_{\mathrm{ref}}=593$. The best evolved task agent shown below achieves $s_{\mathrm{raw}}=582$, corresponding to $s_{\mathrm{norm}}=582/593=0.9815$.

\begin{lstlisting}[language=Python, caption={Best evolved task program for Kissing Number (KN)}, label={lst:kn_best}]
# EVOLVE-BLOCK-START
import numpy as np
import itertools

def kissing_number11() -> np.ndarray:
    """
    Constructs a set of points in Z^11 maximizing cardinality such that max norm <= min pairwise distance.
    Uses the M^2=4 shell strategy with a robust Maximum Weight Independent Set (MWIS) solver.
    Target: Beat 593 points.
    """
    d = 11


    # Generate all sign-complete groups for norm^2 <= 4.
    # Group compatibility (all points in G1 compatible with all points in G2) 
    # is equivalent to: dot(|v1|, |v2|) <= (norm1^2 + norm2^2 - 4) // 2.
    gv, gs, gn = [], [], []    

    # Weights 1-4 for values (+-1)
    for w in range(1, 5):
        for c in itertools.combinations(range(d), w):
            v = np.zeros(d, dtype=np.int32)
            v[list(c)] = 1
            gv.append(v)
            gs.append(2**w)
            gn.append(w)

    # Weight 1 for values (+-2)
    for i in range(d):
        v = np.zeros(d, dtype=np.int32)
        v[i] = 2
        gv.append(v)
        gs.append(2)
        gn.append(4)

    gv = np.array(gv)
    gs = np.array(gs, dtype=np.int32)
    gn = np.array(gn, dtype=np.int32)
    num_g = len(gv)

    # Precompute compatibility (adj) and conflict matrices
    # Condition: dot(p, q) <= (||p||^2 + ||q||^2 - 4) // 2
    dots = gv @ gv.T
    norms_sum = gn[:, None] + gn[None, :]
    adj = dots <= (norms_sum - 4) // 2
    conflicts = ~adj
    np.fill_diagonal(conflicts, False)

    # Metrics for heuristics
    deg = conflicts.sum(axis=1)
    wdeg = conflicts @ gs

    # MWIS solver with Migrant Diversification (Configuration Pool Seeding)
    best_sel, best_count, pool = np.zeros(num_g, dtype=bool), 0, []
    import time
    start_time, node_bias = time.time(), np.zeros(num_g, dtype=float)
    np.fill_diagonal(adj, True)
    for seed in range(42, 100000):
        if time.time() - start_time > 280: break
        np.random.seed(seed)

        # 1. Randomized Greedy Phase with Migrant logic (Seed search from mutated good solutions)
        if pool and np.random.rand() < 0.25:
            sel = pool[np.random.randint(len(pool))].copy()
            active = np.where(sel)[0]
            if len(active) > 0: sel[np.random.choice(active, max(1, len(active)//7), replace=False)] = False
            compat = np.ones(num_g, dtype=bool)
            for idx in np.where(sel)[0]: compat &= adj[idx]
        else:
            sel, compat = np.zeros(num_g, dtype=bool), np.ones(num_g, dtype=bool)

        # Expanded Heuristics: Incorporate power-weightings for diversity
        h_choice = seed % 8
        if h_choice == 0: h = gs / (wdeg + 1.1)
        elif h_choice == 1: h = (gs**1.3) / (wdeg + 1.1)
        elif h_choice == 2: h = gs / (deg + 1.1)
        elif h_choice == 3: h = (gs**1.3) / (deg + 1.1)
        elif h_choice == 4: h = np.log2(gs + 1.0) / (deg + 1.1)
        elif h_choice == 5: h = gs / (np.sqrt(wdeg) + 1.1)
        elif h_choice == 6: h = (gs**1.5) / (wdeg + 1.1)
        else: h = np.random.rand(num_g)
        score = h * (0.7 + 0.6 * np.random.rand(num_g)) + node_bias
        order = np.argsort(-score)

        for idx in order:
            if compat[idx]:
                sel[idx] = True; compat &= adj[idx]

        # 2. Local Search Phase (1-Swap with noise-assisted sideways moves)
        for _ in range(4):
            improved = False
            for i in np.random.permutation(num_g):
                confs = np.where(conflicts[i] & sel)[0]
                w_confs = gs[confs].sum()
                if not sel[i]:
                    if gs[i] > w_confs:
                        sel[confs], sel[i] = False, True; improved = True
                    elif gs[i] == w_confs and (len(confs) > 1 or np.random.rand() < 0.1):
                        # Noise-assisted sideways swap helps escape local optima
                        sel[confs], sel[i] = False, True; improved = True
            if not improved: break

        # 3. Evaluation, Reinforcement, and Pool Management
        current_count = gs[sel].sum()
        if current_count > best_count:
            best_count, best_sel = current_count, sel.copy()
            node_bias[sel] += 0.1 # Stronger reinforcement for new global best
            if best_count >= 640: break
        if current_count >= 560: # Threshold to add configuration to the Migrant Pool
            if not any(np.array_equal(sel, p) for p in pool):
                pool.append(sel.copy())
                if len(pool) > 20: pool.pop(0) # Maintain a diverse, sliding pool
    # Reconstruct 11D points from selected sign-complete groups
    final_points = []
    for idx in np.where(best_sel)[0]:
        v = gv[idx]
        supp = np.where(v != 0)[0]
        vals = v[supp]
        for signs in itertools.product([1, -1], repeat=len(supp)):
            p = np.zeros(d, dtype=np.int64)
            p[supp] = np.array(signs) * vals
            final_points.append(p)

    return np.array(final_points)

# EVOLVE-BLOCK-END
\end{lstlisting}

\subsection{Circle Packing (CP)}
For Circle Packing, the goal is to place $n=26$ circles inside the unit square $[0,1]^2$. A candidate solution specifies centers $c_i=(x_i,y_i)$ and radii $r_i$ for $i=1,\dots,n$. Validity requires that every circle lies inside the square,
\begin{equation}
r_i \le x_i \le 1-r_i,\qquad
r_i \le y_i \le 1-r_i,
\end{equation}
and that circles do not overlap,
\begin{equation}
\|c_i-c_j\|_2 \ge r_i+r_j,\qquad i\neq j.
\end{equation}
The evaluator returns the total packed radius,
\begin{equation}
s_{\mathrm{raw}} = \sum_{i=1}^{26} r_i.
\end{equation}
The reference value used for normalization is $s_{\mathrm{ref}}=2.6350$. The best evolved task agent shown below achieves $s_{\mathrm{raw}}=2.6352223118$, corresponding to $s_{\mathrm{norm}}=1.0001$.

\begin{lstlisting}[language=Python, caption={Best evolved task program for Circle Packing (CP)}, label={lst:cp_best}]
# EVOLVE-BLOCK-START
"""Constructor-based circle packing for n=26 circles"""
import numpy as np

def construct_packing():
    """
    Construct a specific arrangement of 26 circles in a unit square
    that attempts to maximize the sum of their radii.
    Returns:
        Tuple of (centers, radii, sum_of_radii)
        centers: np.array of shape (26, 2) with (x, y) coordinates
        radii: np.array of shape (26) with radius of each circle
        sum_of_radii: Sum of all radii
    """
    # Initialize arrays for 26 circles
    n = 26
    centers = np.zeros((n, 2))


    # Initial setup: 5-5-6-5-5 staggered grid for n=26
    rng = np.random.RandomState(42)
    centers_list = []
    rows = [5, 5, 6, 5, 5]
    for i, row_n in enumerate(rows):
        for j in range(row_n):
            centers_list.append([
                (j + 0.5) / row_n + rng.uniform(-0.002, 0.002),
                (i + 0.5) / len(rows) + rng.uniform(-0.002, 0.002)
            ])
    centers = np.clip(np.array(centers_list), 0, 1)
    # Dynamic order: Prioritize boundary circles. 
    # Corner distance uses sqrt (Euclidean) for more stable weighting.
    def get_order(pts):
        d_edge = np.min(np.hstack([pts, 1 - pts]), axis=1)
        corners = np.array([[0, 0], [1, 0], [0, 1], [1, 1]])
        d_corner = np.min(np.sqrt(np.sum((pts[:, np.newaxis, :] - corners)**2, axis=2)), axis=1)
        return np.argsort(d_edge + 0.07 * d_corner)

    def calculate_radii_from(pts, start_k, current_r, current_order):
        """Vectorized greedy radius calculation with numerical stability via hypot."""
        r = current_r.copy()
        px, py = pts[:, 0], pts[:, 1]
        bounds = np.minimum(np.minimum(px, 1.0 - px), np.minimum(py, 1.0 - py))
        for k in range(start_k, n):
            idx = current_order[k]
            ri = bounds[idx]
            if k > 0:
                prev = current_order[:k]
                # Hypot is more stable for high-precision coordinate descent
                dists = np.hypot(px[idx] - px[prev], py[idx] - py[prev])
                ri = min(ri, np.min(dists - r[prev]))
            r[idx] = max(0.0, ri)
        return r

    # Initial Solution
    order = get_order(centers)
    order_lookup = {idx: i for i, idx in enumerate(order)}
    best_centers = centers.copy()
    best_radii = calculate_radii_from(best_centers, 0, np.zeros(n), order)
    best_sum = np.sum(best_radii)

    # Refinement: High-iteration Simulated Annealing (140,000 steps)
    num_steps = 140000
    for step in range(num_steps):
        progress = step / num_steps
        if step % 1200 == 0:
            order = get_order(best_centers)
            order_lookup = {idx: i for i, idx in enumerate(order)}
            best_radii = calculate_radii_from(best_centers, 0, np.zeros(n), order)
            best_sum = np.sum(best_radii)

        # Cubic cooling provides a long 'warm' tail for staggered structure refinement
        temp = 0.05 * (1.0 - progress)**3.0
        target = rng.randint(0, n)
        old_pos = best_centers[target].copy()

        # Multi-phase search: Exploration -> Refinement -> Quenching
        if progress < 0.8:
            noise = rng.normal(0, temp + 1e-10, 2) if step % 2 == 0 else rng.uniform(-temp, temp, 2)
            m_scale = 0.08
        elif progress < 0.96:
            noise = rng.normal(0, temp + 1e-11, 2)
            m_scale = 0.015
        else:
            noise = rng.normal(0, 5e-9, 2)
            m_scale = 0.0
        best_centers[target] = np.clip(old_pos + noise, 0, 1)

        k = order_lookup[target]
        new_radii = calculate_radii_from(best_centers, k, best_radii, order)
        new_sum = np.sum(new_radii)
        if new_sum > best_sum or (m_scale > 0 and temp > 1e-9 and rng.rand() < np.exp((new_sum - best_sum) / (temp * m_scale))):
            best_sum, best_radii = new_sum, new_radii
        else:
            best_centers[target] = old_pos

    # 3. Final Multi-Scale Micro-Quenching (High-Precision Coordinate Descent)
    # Systematic search in 16 directions (polar + cardinal) to address hexagonal packing
    angles = np.linspace(0, 2*np.pi, 16, endpoint=False)
    search_dirs = [(np.cos(a), np.sin(a)) for a in angles]
    for scale in [1e-5, 1e-7, 1e-9, 1e-11]:
        improved_scale = True
        while improved_scale:
            improved_scale = False
            for idx in range(n):
                k_start = order_lookup[idx]
                for dx, dy in search_dirs:
                    old_x, old_y = best_centers[idx]
                    best_centers[idx, 0] = np.clip(old_x + dx * scale, 0, 1)
                    best_centers[idx, 1] = np.clip(old_y + dy * scale, 0, 1)
                    nr = calculate_radii_from(best_centers, k_start, best_radii, order)
                    ns = np.sum(nr)
                    if ns > best_sum + 1e-15:
                        best_sum, best_radii, improved_scale = ns, nr, True
                    else:
                        best_centers[idx, 0], best_centers[idx, 1] = old_x, old_y
    return best_centers, best_radii, best_sum

# EVOLVE-BLOCK-END
# This part remains fixed (not evolved)

def run_packing():
    """Run the circle packing constructor for n=26"""
    centers, radii, sum_radii = construct_packing()
    return centers, radii, sum_radii


def visualize(centers, radii):
    """
    Visualize the circle packing

    Args:
        centers: np.array of shape (n, 2) with (x, y) coordinates
        radii: np.array of shape (n) with radius of each circle
    """
    import matplotlib.pyplot as plt
    from matplotlib.patches import Circle
    fig, ax = plt.subplots(figsize=(8, 8))
    # Draw unit square
    ax.set_xlim(0, 1)
    ax.set_ylim(0, 1)
    ax.set_aspect("equal")
    ax.grid(True)
    # Draw circles
    for i, (center, radius) in enumerate(zip(centers, radii)):
        circle = Circle(center, radius, alpha=0.5)
        ax.add_patch(circle)
        ax.text(center[0], center[1], str(i), ha="center", va="center")
    plt.title(f"Circle Packing (n={len(centers)}, sum={sum(radii):.6f})")
    plt.show()

if __name__ == "__main__":
    centers, radii, sum_radii = run_packing()
    print(f"Sum of radii: {sum_radii}")
    # AlphaEvolve improved this to 2.635
    # Uncomment to visualize:
    visualize(centers, radii)
\end{lstlisting}

\subsection{Heilbronn Triangle (HT)}
For Heilbronn Triangle, the task is to place $n=11$ points inside the equilateral triangle
\begin{equation}
\Delta = \mathrm{conv}\left\{(0,0),\,(1,0),\,\left(\tfrac{1}{2},\tfrac{\sqrt{3}}{2}\right)\right\},
\end{equation}
so as to maximize the minimum area among all triangles induced by triples of points. For a valid point set $P=\{p_1,\dots,p_{11}\}\subset \Delta$, the evaluator computes
\begin{equation}
s_{\mathrm{raw}}=\min_{1\le i<j<k\le 11}\mathrm{Area}(p_i,p_j,p_k),
\end{equation}
recorded in our logs as \texttt{min\_area\_normalized}. The reference value used for normalization in the current experiments is $s_{\mathrm{ref}}=0.036529889880030156$. The best evolved task agent shown below achieves $s_{\mathrm{raw}}=0.0365253447400576$, corresponding to $s_{\mathrm{norm}}=0.9999$.
\begin{lstlisting}[language=Python, caption={Best evolved task program for Heilbronn Triangle (HT)}, label={lst:ht_best}]
# EVOLVE-BLOCK-START
import numpy as np

def heilbronn_triangle11() -> np.ndarray:
    """
    Generates an optimized arrangement of 11 points within an equilateral triangle 
    to maximize the area of the smallest triangle formed by any three points.
    Vertices are at (0,0), (1,0), and (0.5, sqrt(3)/2).
    """
    n = 11
    sqrt3 = 1.7320508075688772
    h = 0.8660254037844386
    rng = np.random.RandomState(42)

    # Restore Genetic Key: High-performance vectorized search with refined targeted hill climbing
    tri_idx = np.array([[i, j, k] for i in range(n) for j in range(i + 1, n) for k in range(j + 1, n)])
    i0, i1, i2 = tri_idx.T
    pop_size, steps, rows = 64, 57000, np.arange(64)

    # Archetype 4-3-2-2 row configuration
    base = np.array([[0,0],[.33,0],[.67,0],[1,0],[.15,.25],[.5,.25],
                     [.85,.25],[.3,.5],[.7,.5],[.43,.73],[.57,.73]])
    pts = np.tile(base, (pop_size, 1, 1))
    pts[1:] += rng.uniform(-0.04, 0.04, pts[1:].shape)

    # Inline initial projection and area logic
    pts[:, :, 1] = np.clip(pts[:, :, 1], 0, h)
    pts[:, :, 0] = np.clip(pts[:, :, 0], pts[:, :, 1]/sqrt3, 1.0 - pts[:, :, 1]/sqrt3)

    def calc_areas_pop(p):
        p1, p2, p3 = p[:, i0], p[:, i1], p[:, i2]
        return 0.5 * np.abs((p2[...,0]-p1[...,0])*(p3[...,1]-p1[...,1]) - (p3[...,0]-p1[...,0])*(p2[...,1]-p1[...,1]))

    areas = calc_areas_pop(pts)
    min_areas = np.min(areas, axis=1)

    for step in range(steps):
        amp = 0.12 * (0.99987 ** step)
        selector, pts_to_mut = rng.rand(pop_size), rng.randint(0, n, pop_size)

        # Targeted Mutation: 95% focus on points within the 5 smallest bottleneck triangles
        mask = selector < 0.95
        if np.any(mask):
            num_t = np.sum(mask)
            worst = np.argpartition(areas[mask], 5, axis=1)[:, :5]
            sel_tri = worst[np.arange(num_t), rng.randint(0, 5, size=num_t)]
            pts_to_mut[mask] = tri_idx[sel_tri, rng.randint(0, 3, size=num_t)]
        old_coords = pts[rows, pts_to_mut].copy()
        mut_noise = rng.standard_normal((pop_size, 2)) if step % 2 else rng.uniform(-1, 1, (pop_size, 2))
        pts[rows, pts_to_mut] += mut_noise * amp

        # Inline boundary projection
        py = np.clip(pts[rows, pts_to_mut, 1], 0, h)
        pts[rows, pts_to_mut, 1], pts[rows, pts_to_mut, 0] = py, np.clip(pts[rows, pts_to_mut, 0], py/sqrt3, 1.0 - py/sqrt3)
        new_areas = calc_areas_pop(pts)
        new_min = np.min(new_areas, axis=1)

        # Accept improvements or stability
        keep = new_min >= min_areas
        pts[rows[~keep], pts_to_mut[~keep]] = old_coords[~keep]
        areas[keep], min_areas[keep] = new_areas[keep], new_min[keep]

        # Population Sync: Clones the best configurations to the worst ones to accelerate convergence
        if step > 0 and step % 1000 == 0:
            order = np.argsort(min_areas)
            pts[order[:15]] = pts[order[-15:]].copy() + rng.uniform(-amp*0.15, amp*0.15, (15, n, 2))
            py_s = np.clip(pts[order[:15], :, 1], 0, h)
            pts[order[:15], :, 1], pts[order[:15], :, 0] = py_s, np.clip(pts[order[:15], :, 0], py_s/sqrt3, 1.0 - py_s/sqrt3)
            areas[order[:15]] = calc_areas_pop(pts[order[:15]])
            min_areas[order[:15]] = np.min(areas[order[:15]], axis=1)
    return pts[np.argmax(min_areas)]
# EVOLVE-BLOCK-END
\end{lstlisting}

\end{document}